\documentclass[letterpaper]{article} 
\usepackage{aaai23}  
\usepackage{times}  
\usepackage{helvet}  
\usepackage{courier}  
\usepackage[hyphens]{url}  
\usepackage{graphicx} 
\usepackage{natbib}  
\usepackage{caption} 
\usepackage{algorithm}
\usepackage{algorithmic}

\usepackage{amsmath,amsfonts}
\usepackage{array}
\usepackage[caption=false,font=normalsize,labelfont=sf,textfont=sf]{subfig}
\usepackage{textcomp}
\usepackage{stfloats}
\usepackage{url}
\usepackage{verbatim}
\usepackage{graphicx}
\usepackage{cite}

\usepackage{color}
\usepackage{makecell}
\usepackage{booktabs}
\usepackage{multirow}
\usepackage{array}
\usepackage{pifont}
\usepackage[capitalize]{cleveref}
\usepackage[switch]{lineno}

\usepackage{newfloat}
\usepackage{listings}
\DeclareCaptionStyle{ruled}{labelfont=normalfont,labelsep=colon,strut=off} 
\floatstyle{ruled}
\newfloat{listing}{tb}{lst}{}
\floatname{listing}{Listing}
\title{From Coarse to Fine: Hierarchical Pixel Integration for Lightweight Image Super-Resolution}
\author{
    Jie Liu\equalcontrib,
    Chao Chen\equalcontrib,
    Jie Tang \thanks{Corresponding author.},
    Gangshan Wu
}
\affiliations{

    State Key Laboratory for Novel Software Technology, Nanjing University, China\\
    liujie@nju.edu.cn, chenchao@smail.nju.edu.cn, \{tangjie,gswu\}@nju.edu.cn
}

\usepackage{bibentry}

\begin{document}

\maketitle

\begin{abstract}
Image super-resolution (SR) serves as a fundamental tool for the processing and transmission of multimedia data. Recently, Transformer-based models have achieved competitive performances in image SR. They divide images into fixed-size patches and apply self-attention on these patches to model long-range dependencies among pixels. However, this architecture design is originated for high-level vision tasks, which lacks design guideline from SR knowledge. In this paper, we aim to design a new attention block whose insights are from the interpretation of Local Attribution Map (LAM) for SR networks.  Specifically, LAM presents a hierarchical importance map where the most important pixels are located in a fine area of a patch and some less important pixels are spread in a coarse area of the whole image. To access pixels in the coarse area, instead of using a very large patch size, we propose a lightweight Global Pixel Access (GPA) module that applies cross-attention with the most similar patch in an image. In the fine area, we use an Intra-Patch Self-Attention (IPSA) module to model long-range pixel dependencies in a local patch, and then a $3\times3$ convolution is applied to process the finest details. In addition, a Cascaded Patch Division (CPD) strategy is proposed to enhance perceptual quality of recovered images. Extensive experiments suggest that our method outperforms state-of-the-art lightweight SR methods by a large margin. Code is available at \url{https://github.com/passerer/HPINet}.
\end{abstract}

\section{Introduction}

Single-Image Super-Resolution (SISR) aims to recover a visually pleasing high-Resolution (HR) image from its Low-Resolution (LR) counterpart. SISR is widely used in many multimedia applications such as facial recognition on low-resolution images and server costs reduction for media transmission. With the success of Convolutional Neural Networks (CNNs), CNN-based models~\cite{ecbsr,fsrcnn,memnet,srmd,srfbn,edsr,rdn} have become the mainstream of SISR due to the natural local processing ability of convolution kernels.

\begin{figure}[t]
\centering
\includegraphics[width=0.9\linewidth]{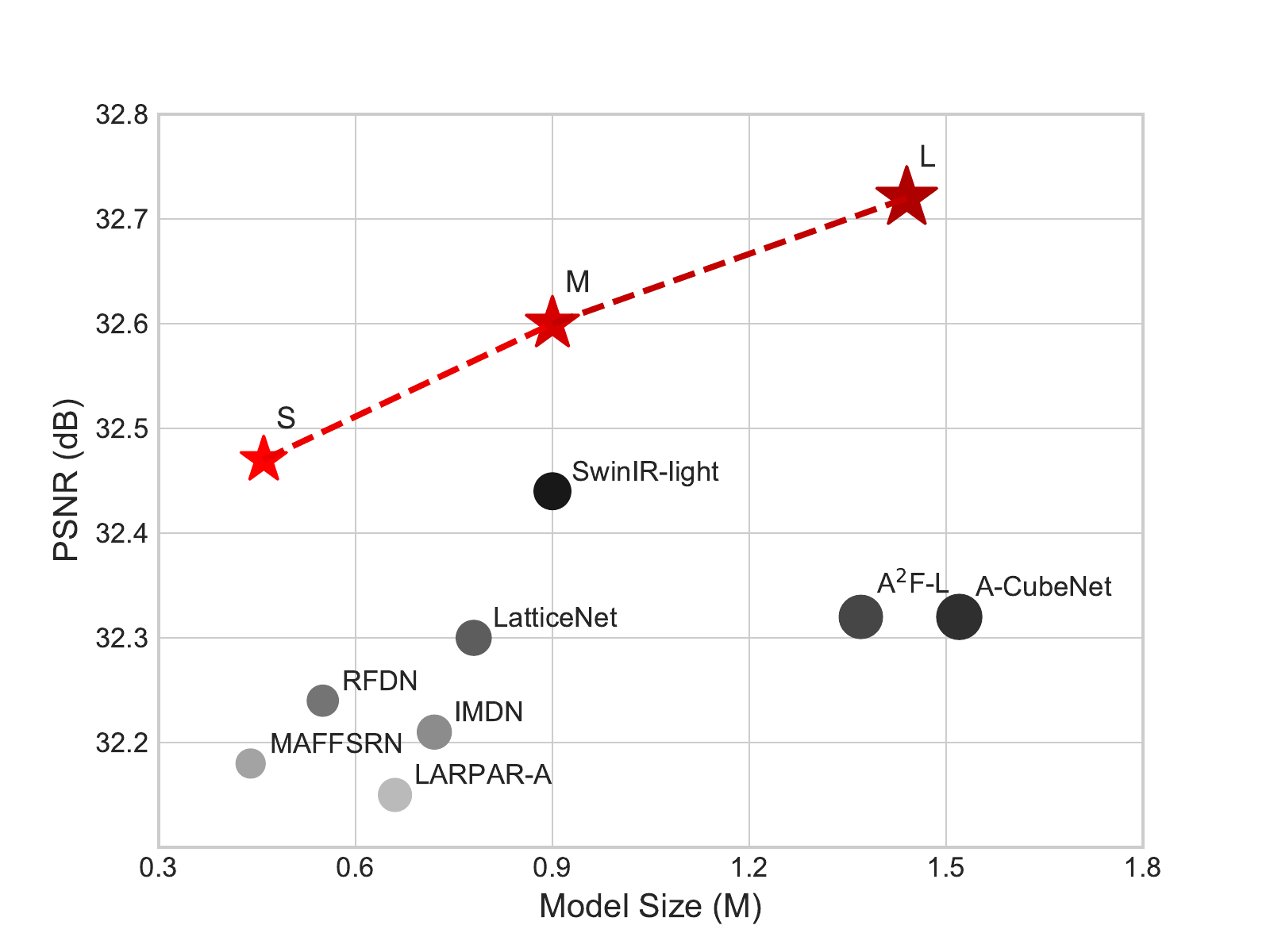}
\caption{PSNR \textit{v.s.} model size for $\times 4$ SR on Set5. We compare our HPINet-S/M/L with other lightweight SR models of various sizes, including MAFFSRN~\cite{maffsrn}, RFDN~\cite{rfdn}, LAPAR-A \cite{lapar}, IMDN~\cite{imdn}, LatticeNet~\cite{latticenet}, SwinIR-light~\cite{swinir}, A$^2$F-L~\cite{a2f} and A-cubeNet~\cite{acubenet}.}\label{fig:psnr_size}
\end{figure}

Though CNN-based models have achieved significant improvement compared with traditional methods, they suffer from some inherent problems brought by the local processing principle of convolution kernels. Recent study~\cite{LAM} shows that SR networks with a wide range of involved input pixels could achieve better performance. However, most of the CNN-based models adopt a small convolution kernel (\textit{e.g.} $3\times 3$) where only a limited range of pixels are aggregated. 
Alternatively, the self-attention mechanism in Transformer~\cite{transformer} models can model long-range dependencies of input pixels and several attempts have already made in the task of SISR. In this paper, we also focus on the self-attention mechanism for aggregating pixels in a wide range of input pixels.

\begin{figure}[t]
\centering
\includegraphics[width=0.9\linewidth]{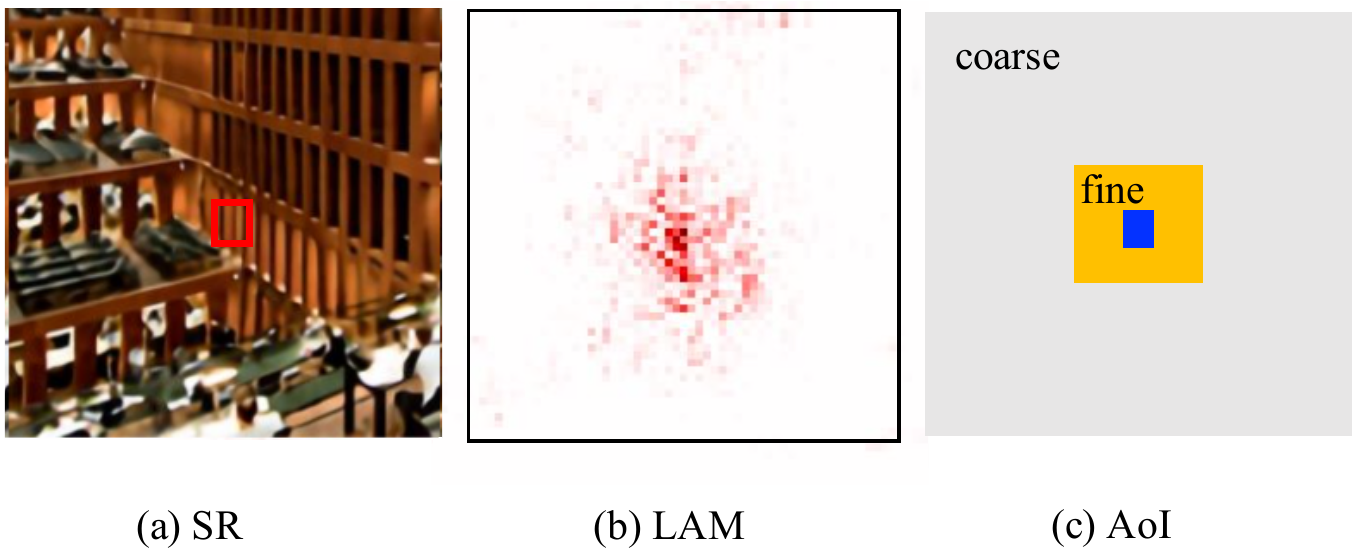}
\caption{(a) SR result of RNAN~\cite{rnan}. (b) Local Attribution Map (LAM)~\cite{LAM}. (c) Area of Interest (AoI). The AoI is defined according 
    to the coarse-to-fine hierarchy of important pixels in LAM. 
}\label{fig:LAM}
\end{figure}

The pioneering vision Transformer  \cite{transformer} adopts a redundant attention manner, whose computation complexity is quadratic to the image size. The large computation complexity makes it difficult to be applied for high-resolution predictions in SISR task. Some recent proposals apply self-attention within a small spatial region \cite{ipt, swinir,uformer} to alleviate this issue. However, these methods lack direct interaction between distant pixels, which is critical for achieving good performance~\cite{LAM}. In~\cite{LAM}, a Local Attribution Map (LAM) is proposed to give a deep understanding of SR networks. As shown in \cref{fig:LAM}b, the LAM map represents the importance of each pixel in the input LR image with respect to the SR of the patch marked with a red box (\cref{fig:LAM}a). Based on the LAM map, we define a Area of Interest (AoI) map in \cref{fig:LAM}c.
The AoI map presents a hierarchical manner that the most important pixels are located in a fine area (yellow area in \cref{fig:LAM}c) of a patch and some less important pixels are spread in a coarse area (gray area in \cref{fig:LAM}c) of the whole image. Further, we use a blue box in \cref{fig:LAM}c to represent the area with highest interest. In this paper, we use the terms coarse and fine to describe the density of important pixels and the granularity of operations interchangeably.
According to this hierarchical partition, we believe that a well-designed SR network should (1) have the ability to focus on processing the finest detail in very local area (\textit{e.g.,} the blue area); (2) be capable of modeling long-range dependencies in a certain area (\textit{e.g.,} the yellow area) to utilize the surrounding context for better reconstruction; (3) have a mechanism to access important pixels in a coarse area (\textit{e.g.,} the gray area) of an input image.   

Based on these observations, we propose a novel Hierarchical Pixel Integration (HPI) block that consists of three main parts: a Global Pixel Access (GPA)
module, an Intra-Patch Self-Attention (IPSA) module, and a $3\times3$ convolution. The GPA module is responsible for pixel access in coarse area.
Specifically, a similarity map between each pair of image patches is calculated and the most similar patch is selected to conduct
cross-attention with current patch. In this way, important pixels in the coarse area can be integrated into current patch efficiently.
After fusing pixels from coarse area, we apply the standard self-attention to model long-range dependencies in the fine area of a patch. Finally,
a $3\times 3$ convolution is adopted to refine local details in the finest area.
Besides, we found in experiments that the vanilla patch division hinders the perceptual quality of recovered images. To tackle this issue, we propose
to use a Cascaded Patch Division (CPD) that gradually enlarges the patch window in different blocks.

In summary, our main contributions are as follows:
\begin{itemize}
    \item We introduce a hierarchical interpretation of the Local Attention Map (LAM)~\cite{LAM} and devise a new attention block for image SR.

    \item Instead of the vanilla patch division method that fixes the patch size throughout the network, a Cascaded Patch Division (CPD) strategy is applied for better perceptual quality in terms of LPIPS.

    \item We propose a lightweight Hierarchical Pixel Access Network (HPINet) that outperforms existing lightweight SR methods by a large margin. 
        Most notably, our HPINet achieves results that match state-of-the-art large (around 15M) SR models using less than 1.5M parameters.
\end{itemize}

\section{RELATED WORK}
\subsection{Deep Neural Networks for SR}
Deep neural networks have become the most popular methodology for image SR in the past several years thanks to their great representation power. Since   \cite{srcnn} first uses three convolution layers to map the low-resolution images to high-resolution images, various CNN-based networks have further enhanced the state-of-the-art performance via better architecture design, such as residual connections  \cite{vsdr,drcn,rfdn,rfanet,esrgan}, U-shape architecture \cite{encoderdecoder,verydeepencoder,waveletcnn} and attention mechanisms \cite{imdn,rcan,pan,han,mgan}. 

In addition to improving accuracy,  some CNN-based works pursue a lightweight and economical structure. Specifically, IDN \cite{idn} stacks multiple information distillation blocks to extract more useful information. Each block divides feature maps into two parts. One part is reserved and the other is further enhanced by several convolutions. After that two kinds of features are combined for more abundant information. Based on information distillation block, IMDN \cite{imdn} utilizes information multi-distillation block, which retains and process partial feature maps step-by-step, and aggregates them by contrast-aware channel attention mechanism. Furthermore, RFDN \cite{rfdn} applies intensive residual learning to distill more efficient feature representations.

While CNN-based methods have dominated this field for a long time, recent works introduce Transformer~\cite{transformer} and make impressive progress. IPT \cite{ipt} is one of the first methods use Transformer in image SR. However, IPT contains a massive amount of parameters and requires large-scale datasets for training, which limits its practice in real applications. Modified from Swin Transformer \cite{swintransformer}, SwinIR \cite{swinir} limits the attention region in fixed-size windows and uses shift operation to exchange information through nearby windows. As such, it achieves a better trade-off between PSNR and the number of parameters than prevalent CNN-based models. Despite the success, SwinIR set the window size as $8\times 8$, which is larger than a regular convolution (often $3\times 3$ or $5\times 5$) but much smaller than the conventional Transformer (i.e., the entire image), thereby requiring stacking a great number of blocks to achieve sufficiently large receptive fields. In contrast, our HPINet is able to exploit global context even in the first block, which is beneficial for integrating important pixels with fine detail in the early stage.




\begin{figure}[t]
\centering
\includegraphics[width=\linewidth]{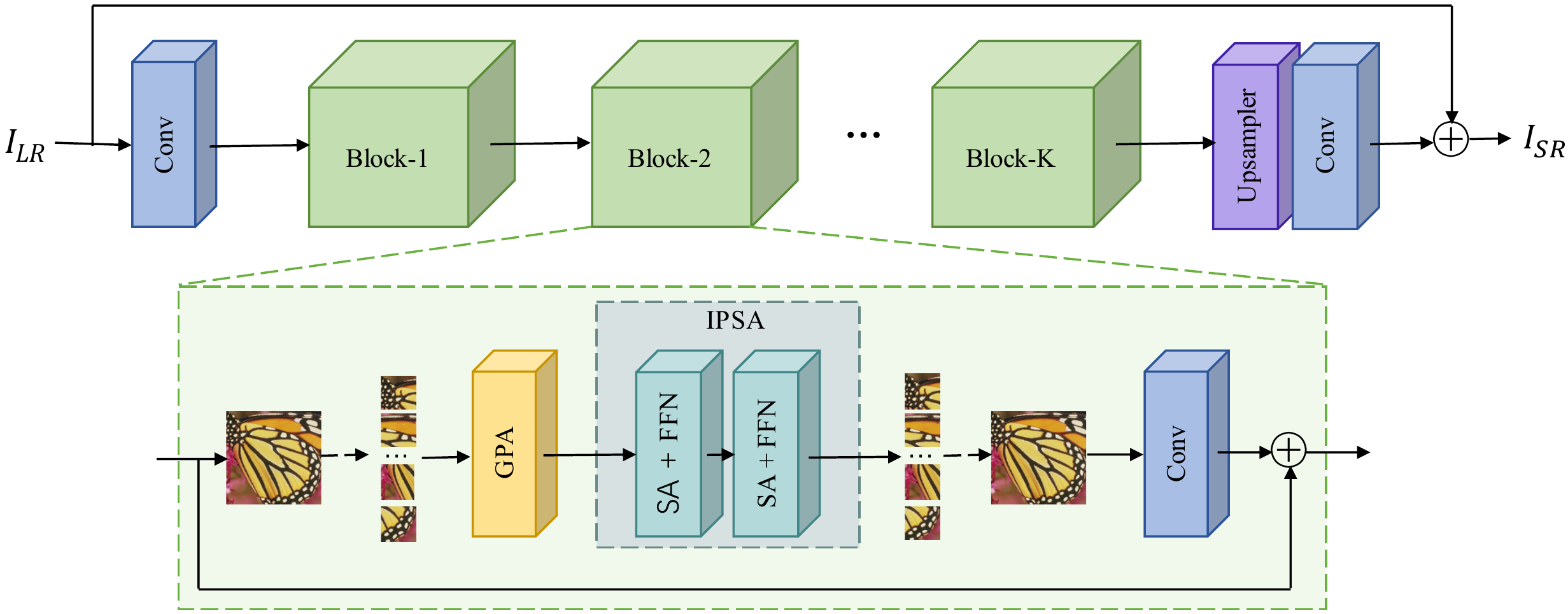}
\caption{Overall network architecture of HPINet. The upsampling operation on skip connection between $I_{LR}$ and $I_{SR}$ is omitted for clarity.
}
\label{fig:overall}
\end{figure}

\begin{figure}[t]
\centering
\subfloat[]{
     \centering
     \includegraphics[width=0.3\linewidth]{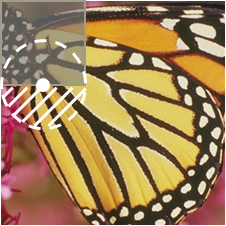}
     \label{fig:division_a}}
    \subfloat[]{
     \centering
     \includegraphics[width=0.3\linewidth]{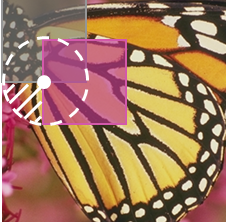}
     \label{fig:division_b}}
    \subfloat[]{
     \centering
     \includegraphics[width=0.3\linewidth]{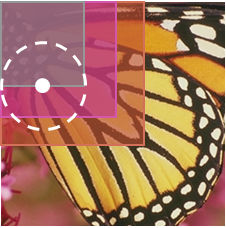}\label{fig:division_c}}
    \caption{Different strategies of patch division. Areas of different colors represent patches in different network layers. The white point represents the sampled pixel whose blind spot is covered with white shadows. (a) vanilla patch division; (b) shift-based patch division; (c) cascaded patch division (ours).}
\label{fig:division}
\end{figure}

\section{Method}
\subsection{Network Architecture}
As shown in \cref{fig:overall}, the proposed HPINet consists of three components: encoder, Hierarchical Pixel Integration (HPI) block, and decoder. Encoder is a $3\times 3$ convolutional layer which serves to map the input image space to a higher dimensional feature space. Let $I_{\text{LR}}$ and $I_{\text{SR}}$ denote the low resolution input image and the super-resolved image, respectively. We first get shallow feature $x_0$  by
\begin{linenomath}
\begin{align}
    x_0 = f_{\text{Encoder}}(I_{\text{LR}}),
\end{align}
\end{linenomath}
where $f_{\text{Encoder}}$ denotes the function of the encoder.
Then deeper features are extracted by $K$ sequential HPI blocks. The HPI block consists of five parts: Cascaded Patch Division (CPD), Global Pixel Access (GPA), Intra-Patch Self-Attention (IPSA), Patch Aggregation (PA), and $3\times3$ Convolution (Conv). The input feature is divided into patches by the CPD module and finally aggregated together via patch aggregation. Formally, in the $i_{\text{th}}$ block, the output feature $x_i$ is obtained by
\begin{linenomath}
\begin{align}
    x_i = x_{i-1} + f_{\text{Conv}}(f_{\text{PA}}(f_{\text{IPSA}}(f_{\text{GPA}}(f_{\text{CPD}}(x_{i-1}))))),
\end{align}
\end{linenomath}
where $f(\cdot)$ denotes the function of each individual component. As with existing works, residual learning is used to assist the training process.

Finally, decoder with pixel-shuffle operations \cite{pixelshuffle} is adopted to get the global residual information, which is added to $I_{\text{LR}}$ for restoring the high-resolution output
\begin{linenomath}
\begin{align}
    I_{\text{SR}} = I_{\text{LR}} + f_{\text{Decoder}}(x_{K}).
\end{align}
\end{linenomath}

\begin{figure*}[t]
\centering
\includegraphics[width=0.96\linewidth]{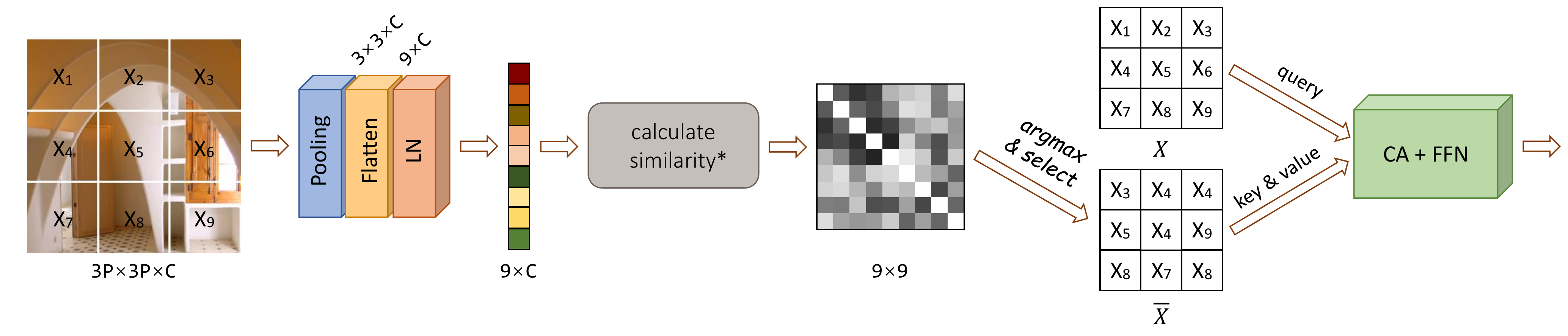}
\caption{Illustration of Global Pixel Access (GPA) module. * denotes that diagonal elements of the similarity map are set to zero.}
\label{fig:GPA}
\end{figure*}

\subsection{Cascaded Patch Division}\label{subsection:hpd}

In order to reduce the computational complexity for lightweight image SR, we split the input feature map into a collection of equal-sized patches and process each patch independently. Specifically, given an input $\mathbf{X}$ of size $H\times W\times C$, we split it into a set of square patches, \textit{i.e.}, $\mathbf{X} = \{\mathbf{X}_i\in \mathbb{R}^{P^2\times C} \:|\: i=1,\cdots,N \}$, where $P$ is the patch size and $N$ stands for the total number of patches. Each patch must satisfy
\begin{linenomath}
\begin{align}
  u \:\%\: P &= 0 \quad \text{or} \quad u=H-P\; \\
  v \:\%\: P &= 0 \quad \text{or} \quad v=W-P\;
\end{align}
\end{linenomath}
where $(u,v)$ denotes the coordinate of pixel in the top-left corner. All patches can be processed in parallel, after which the outputs are pasted to their original location in patch aggregation module. It is worthwhile to note that such cropping strategy is adaptive to arbitrary input size, which means no padding pixels are needed.

Though this vanilla division greatly reduces computational cost, it limits the receptive field of boundary pixels. Unlike pixels around the center of a patch, border pixels can not directly interact with neighboring pixels that are out of the patch (see \cref{fig:division}a), which may deteriorate the visual quality of reproduced images. In practice, patch overlapping is useful but not effective enough. Other attempts like \cite{swinir} use shift operations to re-divide patches, but there are still pixels that fail to reach their neighbors directly (see \cref{fig:division}b). Inspired by the idea of progressive learning, we just assign different patch size $P$ to different blocks in a cascaded manner. In other words, we enlarge $P$ progressively in the network. As a result, the border pixels of a block could be in the center of later blocks. Therefore, there exist non-persistent boundaries (see \cref{fig:division}c). It also comes with the added bonus that a smaller receptive filed in the shallow layers helps stabilize the training process, while a larger receptive field in deep layers enables smoother pixel integration. Experiments indicate that the Cascaded Patch Division (CPD) strategy is fairly simple yet effective.

\subsection{Intra-Patch Self-Attention}
Before the introduction of our GPA module, we first describe the detail of Intra-Patch Self-Attention (IPSA) for a better understanding.
IPSA follows the standard self-attention paradigm \cite{transformer}, whereas there are two changes. Firstly, IPSA is performed at patch level instead of image level. Secondly, positional embedding is removed because of the introduce of the convolutional layer, which can learn positional relations implicitly and make the network more concise and efficient. IPSA is responsible for modeling long-range dependencies in
a patch so that the context information could be fully utilized.

More specifically, for a patch feature $\mathbf{X}\in \mathbb{R}^{P^2 \times C}$, the \textit{query}, \textit{key} and \textit{value} matrices $\mathbf{Q}\in \mathbb{R}^{P^2 \times d}$, $\mathbf{K}\in \mathbb{R}^{P^2 \times d}$, $\mathbf{V}\in \mathbb{R}^{P^2 \times C}$ are computed as 
\begin{linenomath}
\begin{align}\label{eq:qkv}
    \mathbf{Q} = \mathbf{XW}_Q, \quad \mathbf{K}=\mathbf{XW}_K, \quad \mathbf{V}=\mathbf{XW}_V,
\end{align}
\end{linenomath}
where $\mathbf{W}_Q$, $\mathbf{W}_K$ and $\mathbf{W}_V$ are weight matrices that are shared across patches. By comparing the similarity between $\mathbf{Q}$ and $\mathbf{K}$, we obtain a attention map of size $\mathbb{R}^{P^2\times P^2}$ and multiply it with $\mathbf{V}$. Overall, the calculation of Self-Attention (SA) can be formulated as
\begin{linenomath}
\begin{align}
    SA(\mathbf{X}) = \text{softmax}(\mathbf{QK}^T/\sqrt{d})\mathbf{V}.
\end{align}
\end{linenomath}
Here $\sqrt{d}$ is used to control the magnitude of $\mathbf{QK}^T$ before applying the softmax function.
 
Similar to the conventional Transformer layer \cite{transformer}, the Feed Forward Network (FFN) is employed after SA module to further transform features. FFN contains two fully-connected layers, and one GELU non-linearity is applied after the first linear layer. Besides, layer normalization is added before SA module and FFN module, and residual shortcuts after both modules are added as well.

\subsection{Global Pixel Access}
In this part, we aim to integrate important pixels belonging to the coarse area of a LAM map (see \cref{fig:LAM}).
SR Networks with a wider range of effective receptive field have been proven to achieve better performance \cite{LAM}. The problem is how to make the network be capable of modeling global connectivity while maintaining computational efficiency. Since the partition of patch is fixed for a layer, there is no direct connection across patches. A straightforward way is to mix the information of every patch pair exhaustively. However, it is unnecessary and inefficient given the fact that many patches are irrelevant and uninformative. Moreover, redundant interaction may introduce extra noise that hinders the model performance.

Based on these observations, we propose an innovative Global Pixel Access (GPA) module, in which each patch performs adaptive pixel integration with the most correlative counterpart (see \cref{fig:GPA}). To be specific, firstly, all patches are spatially pooled into one-dimensional tokens. These tokens encode the characteristic of the patches, which are later used for similarity calculation and patch matching. This process can be expressed as
\begin{linenomath}
\begin{align}
    \overline{\mathbf{X}}_i = \mathop{\arg\max}\limits_{\mathbf{X}_j}\;L(\mathbf{X}_i)^T L(\mathbf{X}_j),\quad j\neq i,
\end{align}
\end{linenomath}
where $\overline{\mathbf{X}}_i$ is the best-matching patch with $X_i$, and $L(\cdot)$ is the average pooling function along spatial dimension followed by flatten operation and layer normalization. Since the $argmax$ opration is non-differentiable, we replace it with Gumbel-Softmax opration \cite{gumblesoftmax} during training so as to make it possible to train end-to-end. After that, pixel information of $\overline{\mathbf X}_i$ are fused into $\mathbf X_i$ via Cross-Attention (CA) 
\begin{linenomath}
\begin{align}
\mathbf{X}_i = \text{CA}(\mathbf{X}_i, \overline{\mathbf{X}}_i).
\end{align}
\end{linenomath}
As illustrated in \cref{fig:GPA}, CA works in a  similar way to the standard self-attention~\cite{transformer}, but the \textit{key} and \textit{value} are calculated using $\overline{\mathbf X}_i$ in Equation (\cref{eq:qkv}). As a result, GPA can enable global pixel integration while introducing little computational overhead.

\subsection{Local $3\times3$ Convolution}
The aforementioned GPA and IPSA can integrate information from a wide range of pixels. However, as indicates by the LAM map in \cref{fig:LAM}b, the most import pixels usually locate in a local area of a patch. Therefore, we need a more local operation to process local detail in a fine-grained manner. We found that a simple
$3\times 3$ convolution is efficient and effective for this task due to the local processing principle of convolution kernel.

\section{Experiments}
\subsection{Datasets and Evaluation Metrics}
The model is trained with a high-quality dataset DIV2K \cite{div2k}, which is widely used for image SR task. It includes 800 training images together with 100 validation images. Besides, we evaluate our model on five public SR benchmark datasets: Set5 \cite{set5}, Set14
 \cite{set14}, B100 \cite{b100}, Urban100 \cite{urban100} and Manga109 \cite{manga109}.

 We use objective criteria, \textit{i.e.}, peak signal-to-noise ratio (PSNR), structural
similarity index (SSIM) to evaluate our model performance. As adopted in the previous work, the two metrics are both calculated on the Y channel after converting to YCbCr space. Besides, we report total number of parameters to compare the complexity of different models.

\subsection{Implementation Details}\label{sec:implement}
Motivated by \cite{restormer}, we adopt progressive learning during training process. 
The cropped HR image size is initialized as 196$\times$196 and increases to 896$\times$896 epoch by epoch, and batch size is set as 6.
Training images are augmented by random flipping and rotation. All models are trained using Adam algorithm with L1 loss. The learning rate is initialized as $3\times 10^{-4}$ and halved per 200 epochs. For the proposed HPINet, the number of blocks is set as 8 and the corresponding patch size is set as $\left\{12,16,20,24,12,16,20,24\right\}$.  The whole process is implemented by Pytorch on NVIDIA Tesla V100 GPUs. More
specific details will be explained in the respective subsection.

\begin{figure}
\centering
 
     \includegraphics[width=\linewidth]{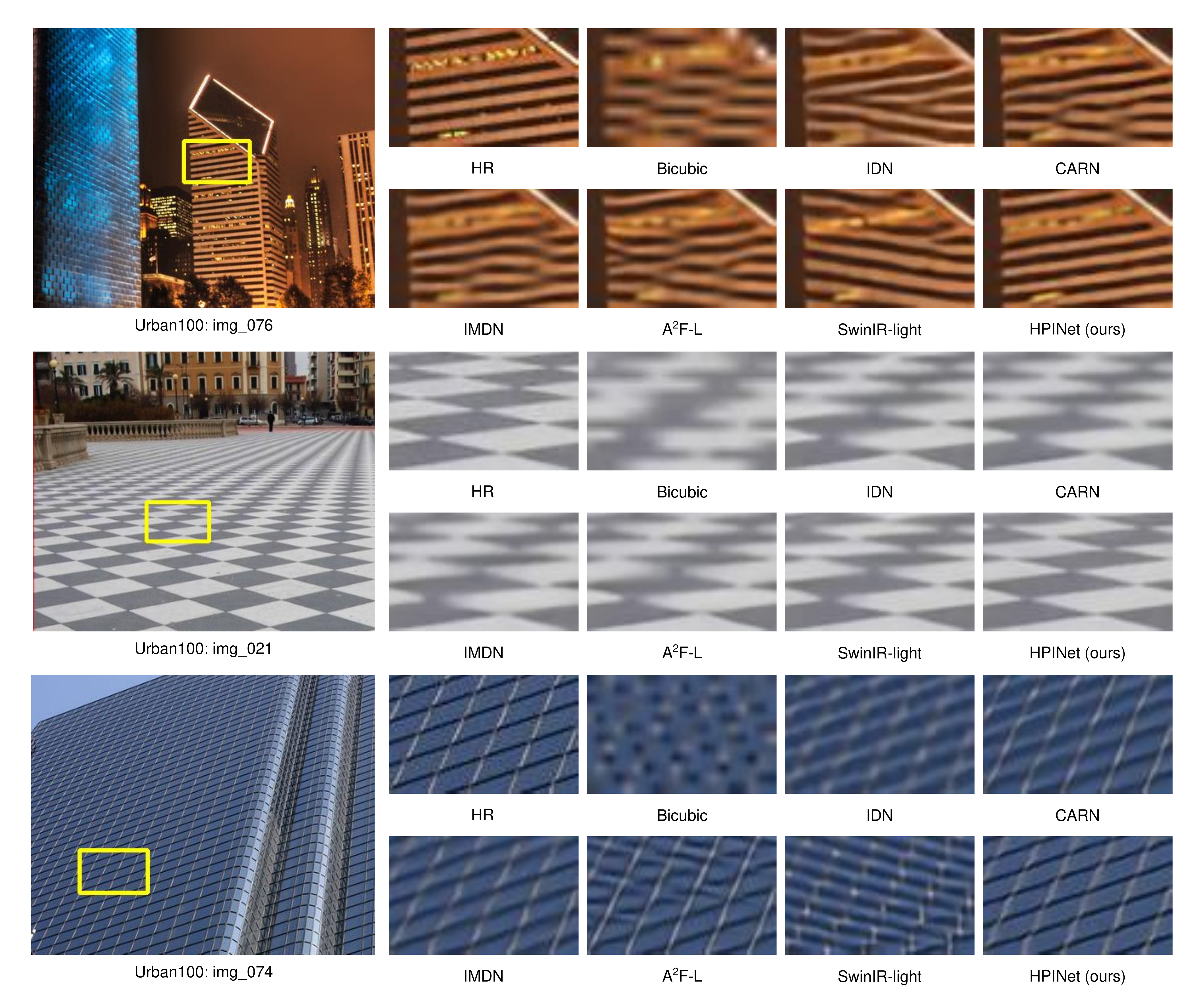}

\caption{Visual comparison with state-of-the-art lightweight SR methods on Urban100, including IDN~\cite{idn}, CARN~\cite{carn}, IMDN~\cite{imdn}, A$^2$F-L~\cite{a2f}, and SwinIR-light~\cite{swinir}.
}\label{fig:visual_results1}
\end{figure}

\begin{table}[t]
\begin{center}
\caption{Quantitative Comparisons of HPINet with Existing Lightweight SR Models on Benchmark Datasets. Our Models are Marked in \textbf{Bold}. Best and Second Best Performance are Highlighted in \textcolor{red}{Red} and \textcolor{blue}{Blue} Colors, Respectively. ``Param'' Is the Number of Parameters \label{table:sota}}
\resizebox{\linewidth}{!}{
\begin{tabular}{clcccccc}
\toprule\noalign{\smallskip}
Scale  & Model & Param & \makecell[c]{Set5\\PSNR/SSIM} & \makecell[c]{Set14\\PSNR/SSIM} & \makecell[c]{B100\\PSNR/SSIM} & \makecell[c]{Urban100\\PSNR/SSIM} & \makecell[c]{Manga109\\PSNR/SSIM}\\
\noalign{\smallskip}
\midrule
\noalign{\smallskip}
\multirow{16}*{$\times 2$}  & FALSR-C  & 0.41M & 37.66/0.9586 & 33.26/0.9140 & 31.96/0.8965 & 31.24/0.9187 & - \\
~ & DRRN  & 0.30M & 37.74/0.9591 & 33.23/0.9136 & 32.05/0.8973 & 31.23/0.9188 & 37.92/0.9760 \\
~  & CARN  & 1.59M & 37.76/0.9590 & 33.52/0.9166 & 32.09/0.8978 & 31.92/0.9256 & 38.36/0.9765 \\
~  & FALSR-A  & 1.02M & 37.82/0.9595 & 33.55/0.9168 & 32.12/0.8987 & 31.93/0.9256 & - \\
~ & IDN  & 0.55M & 37.83/0.9600 & 33.30/0.9148 & 32.08/0.8985 & 31.27/0.9196 & 38.88/0.9774 \\
~ & A$^2$F-SD  & 0.31M & 37.91/0.9602 & 33.45/0.9164 & 32.08/0.8986 & 31.79/0.9246 & 38.52/0.9767 \\
~  & MAFFSRN  & 0.40M & 37.97/0.9603 & 33.49/0.9170 & 32.14/0.8994 & 31.96/0.9268 & - \\
~  & PAN  & 0.26M & 38.00/0.9605 & 33.59/0.9181 & 32.18/0.8997 & 32.01/0.9273 & 38.70/0.9773 \\
~  & IMDN  & 0.69M & 38.00/0.9605 & 33.63/0.9177 & 32.19/0.8999 & 32.17/0.9283 & 38.01/0.9749 \\
~ & LAPAR-A  & 0.55M & 38.01/0.9605 & 33.62/0.9183 & 32.19/0.8999 & 32.10/0.9283 & 38.67/0.9772 \\
~  & RFDN  & 0.53M & 38.05/0.9606 & 33.68/0.9184 & 32.16/0.8994 & 32.12/0.9278 & 38.88/0.9773 \\
~ & A$^2$F-L  & 1.36M & 38.09/0.9607 & 33.78/0.9192 & 32.23/0.9002 & 32.46/0.9313 & 38.95/0.9772 \\
~ & A-CubeNet  & 1.38M & 38.12/0.9609 & 33.73/0.9191 & {\color{blue}32.26}/0.9007 & 32.39/0.9308 & 38.88/0.9776 \\
~  & LatticeNet  & 0.76M & {\color{red}38.15}/{\color{blue}0.9610} & 33.78/0.9193 & 32.25/0.9005 & 32.43/0.9302 & - \\
~  & SwinIR-light  & 0.88M & {\color{blue}38.14}/{\color{red}0.9611} & {\color{blue}33.86}/{\color{blue}0.9206} & {\color{red}32.31}/{\color{blue}0.9012} & {\color{blue}32.76}/{\color{blue}0.9340} & {\color{red}39.12}/{\color{red}0.9783} \\
~  & \textbf{HPINet (Ours)} & 0.78M & 38.12/0.9605 & {\color{red}33.94}/{\color{red}0.9209} & {\color{red}32.31}/{\color{red}0.9013} & {\color{red}32.85}/{\color{red}0.9346} & {\color{blue}39.08}/{\color{blue}0.9771} \\
\midrule
\multirow{14}*{$\times 3$}  & DRRN  & 0.30M & 34.03/0.9244 & 29.96/0.8349 & 28.95/0.8004 & 27.53/0.8378 & 32.74/0.9390 \\
~ & IDN  & 0.55M & 34.11/0.9253 & 29.99/0.8354 & 28.95/0.8013 & 27.42/0.8359 & 32.71/0.9381 \\
~  & A$^2$F-SD  & 0.32M & 34.23/0.9259 & 30.22/0.8395 & 29.01/0.8028 & 27.91/0.8465 & 33.29/0.9424 \\
~ & CARN  & 1.59M & 34.29/0.9255 & 30.29/0.8407 & 29.06/0.8034 & 28.06/0.8493 & 33.50/0.9440 \\
~  & MAFFSRN  & 0.42M & 34.32/0.9269 & 30.35/0.8429 & 29.09/0.8052 & 28.13/0.8521 & - \\
~ & IMDN  & 0.70M & 34.36/0.9270 & 30.32/0.8417 & 29.09/0.8046 & 28.17/0.8519  & 33.61/0.9445 \\
~ & LAPAR-A  & 0.59M & 34.36/0.9267 & 30.34/0.8421 & 29.11/0.8054 & 28.15/0.8523 & 33.51/0.9441 \\
~   & PAN  & 0.26M & 34.40/0.9272 & 30.36/0.8422 & 29.11/0.8049 & 28.11/0.8511+ &  33.61/0.9448 \\
~ & RFDN  & 0.54M & 34.41/0.9273 & 30.34/0.8420 & 29.09/0.8050 & 28.21/0.8525 & 33.67/0.9449 \\
~  & LatticeNet  & 0.77M & 34.53/0.9281 & 30.39/0.8424  & 29.15/0.8059 & 28.33/0.8538 & - \\
~  & A-CubeNet  & 1.56M & 34.53/0.9281 & 30.45/0.8441 & 29.17/0.8068 & 28.38/0.8568 & 33.90/0.9466 \\
~  & A$^2$F-L  & 1.37M & 34.54/{\color{blue}0.9283} & 30.41/0.8436 & 29.14/0.8062  & 28.40/0.8574 & 33.83/0.9463 \\
~ & SwinIR-light & 0.89M & {\color{blue}34.62}/{\color{red}0.9289} & {\color{blue}30.54}/{\color{blue}0.8463} & {\color{blue}29.20}/{\color{blue}0.8082} & {\color{blue}28.66}/{\color{blue}0.8624} & {\color{blue}33.98}/{\color{blue}0.9478} \\
~  & \textbf{HPINet (Ours)} & 0.92M & {\color{red}34.70}/{\color{red}0.9289} & {\color{red}30.63}/{\color{red}0.8480} & {\color{red}29.26}/{\color{red}0.8104} & {\color{red}28.93}/{\color{red}0.8675} & {\color{red}34.31}/{\color{red}0.9487} \\
\midrule
\multirow{14}*{$\times 4$} & DRRN  & 0.30M & 31.68/0.8888 & 28.21/0.7720 & 27.38/0.7284 & 25.44/0.7638 & 29.46/0.8960 \\
~ & IDN  & 0.55M & 31.82/0.8903 & 28.25/0.7730 & 27.41/0.7297 & 25.41/0.7632 & 29.41/0.8942 \\
~  & A$^2$F-SD  & 0.32M & 32.06/0.8928 & 28.47/0.7790 & 27.48/0.7373 & 25.80/0.7767 & 30.16/0.9038 \\
~ & CARN  & 1.59M & 32.13/0.8937 & 28.60/0.7806 & 27.58/0.7349 & 26.07/0.7837 & 30.47/0.9084 \\
~ & PAN  & 0.27M & 32.13/0.8948 & 28.61/0.7822 & 27.59/0.7363 & 26.11/0.7854 & 30.51/0.9095 \\
~  & LAPAR-A  & 0.66M & 32.15/0.8944 & 28.61/0.7818 & 27.61/0.7366 & 26.14/0.7871 & 30.42/0.9074 \\
~ & MAFFSRN  & 0.44M & 32.18/0.8948 & 28.58/0.7812 & 27.57/0.7361 & 26.04/0.7848 & - \\
~  & IMDN   & 0.72M & 32.21/0.8948 & 28.58/0.7811 & 27.56/0.7353 & 26.04/0.7838 & 30.45/0.9075 \\
~ & RFDN  & 0.55M & 32.24/0.8952 & 28.61/0.7819 & 27.57/0.7360 & 26.11/0.7858 & 30.58/0.9089 \\
~  & LatticeNet & 0.78M & 32.30/0.8962 & 28.68/0.7830  & 27.62/0.7367 & 26.25/0.7873 & -\\
~  & A$^2$F-L  & 1.37M & 32.32/0.8964 & 28.67/0.7839 &  27.62/0.7379 & 26.32/0.7931 & 30.72/0.9115 \\
~  & A-CubeNet & 1.52M & 32.32/0.8969 & 28.72/0.7847 & 27.65/0.7382 & 26.27/0.7913 & 30.81/0.9114\\
~ & SwinIR-light  & 0.90M & {\color{blue}32.44}/{\color{blue}0.8976} & {\color{blue}28.77}/{\color{blue}0.7858} & {\color{blue}27.69}/{\color{blue}0.7406} & {\color{blue}26.47}/{\color{blue}0.7980} & {\color{blue}30.92}/{\color{blue}0.9151} \\
~  & \textbf{HPINet (Ours)} & 0.90M & {\color{red}32.60}/{\color{red}0.8986} & {\color{red}28.87}/{\color{red}0.7874} & {\color{red}27.73}/{\color{red}0.7419} & {\color{red}26.71}/{\color{red}0.8043} & {\color{red}31.19}/{\color{red}0.9161} \\
\bottomrule
\end{tabular}}
\end{center}
\end{table}
\begin{table*}
\begin{center}
    \caption{Ablation Study on the Cascaded Patch Division (CPD) and Global Pixel Access (GPA). Metrics (PSNR$\uparrow$/SSIM$\uparrow$/LPIPS$\downarrow$) Are Calculated on Benchmark Datasets with a Scale Factor of 4, Where ``$\uparrow$'' Indicates Higher Is Better and ``$\downarrow$'' Means Lower Is Better. Best and Second Best Result Are in \textcolor{red}{Red} and \textcolor{blue}{Blue} Colors, Respectively}
\label{tab:ablation}
\resizebox{\linewidth}{!}{
\begin{tabular}{c|cc|ccccccc}
\toprule
Model& CPD & GPA & Param & Multi-Adds & \makecell[c]{Set5\\ \hline PSNR$\uparrow$/SSIM$\uparrow$/LPIPS$\downarrow$} & \makecell[c]{Set14\\ \hline PSNR$\uparrow$/SSIM$\uparrow$/LPIPS$\downarrow$} & \makecell[c]{B100\\ \hline PSNR$\uparrow$/SSIM$\uparrow$/LPIPS$\downarrow$} & \makecell[c]{Urban100\\ \hline PSNR$\uparrow$/SSIM$\uparrow$/LPIPS$\downarrow$} & \makecell[c]{Manga109\\ \hline PSNR$\uparrow$/SSIM$\uparrow$/LPIPS$\downarrow$} \\ 
\midrule
\ding{172}&\ding{55} &\ding{55} & 895K & 121.7G & 32.46/0.8974/0.1760 & 28.83/0.7859/0.2844 & 27.70/0.7400/0.3444 & 26.57/0.7995/0.2125 & 31.00/0.9144/0.1149 \\ 
\ding{173}&\ding{51} & \ding{55} & 895K & 123.9G & \textcolor{blue}{32.50}/\textcolor{blue}{0.8979}/\textcolor{blue}{0.1735} & 28.83/\textcolor{blue}{0.7869}/\textcolor{red}{0.2817} & 27.71/\textcolor{blue}{0.7416}/\textcolor{red}{0.3421} & \textcolor{blue}{26.63}/\textcolor{blue}{0.8018}/\textcolor{red}{0.2094} & 31.04/\textcolor{blue}{0.9151}/\textcolor{blue}{0.1139} \\ 
\ding{174}&\ding{55} &\ding{51} & 896K & 121.8G & \textcolor{blue}{32.50}/0.8977/0.1757 & \textcolor{blue}{28.86}/0.7865/0.2868 & \textcolor{blue}{27.72}/0.7408/0.3476 & \textcolor{blue}{26.63}/0.8006/0.2163 & \textcolor{blue}{31.08}/0.9150/0.1155 \\ 
\ding{175} & \ding{51} & \ding{51} & 896K & 124.0G & \textcolor{red}{32.60}/\textcolor{red}{0.8986}/\textcolor{red}{0.1729} & \textcolor{red}{28.87}/\textcolor{red}{0.7874}/\textcolor{blue}{0.2826} & \textcolor{red}{27.73}/\textcolor{red}{0.7419}/\textcolor{blue}{0.3433} & \textcolor{red}{26.71}/\textcolor{red}{0.8043}/\textcolor{blue}{0.2100} & \textcolor{red}{31.19}/\textcolor{red}{0.9161}/\textcolor{red}{0.1127} \\
\bottomrule
\end{tabular}}
\end{center}
\end{table*}

\subsection{Comparisons with the state-of-the-arts}

We compare the proposed HPINet with commonly used lightweight SR models for
upscaling factor $\times$2, $\times$3, and $\times$4, including FALSR \cite{faslr}, DRRN \cite{drrn}, A$^2$F \cite{a2f}, MAFFSRN \cite{maffsrn}, PAN \cite{pan}, IDN \cite{idn}, IMDN \cite{imdn}, LAPAR \cite{lapar}, RFDN \cite{rfdn}, SwinIR \cite{swinir}, CARN \cite{carn}, LatticeNet \cite{latticenet} and A-CubeNet \cite{acubenet}. The comparison results are classified into several groups according to the upscaling factor.

\subsubsection{Quantitative Comparison}
\cref{table:sota} shows quantitative results in terms of PSNR and SSIM on five benchmark datasets obtained by different algorithms. As shown in \cref{table:sota}, our HPINet achieves the best performance for $\times 3$ and $\times 4$ SR. For $\times 2$ SR, we also achieve superior
results than the state-of-the-art SwinIR-light model on Set14, B100 and Urban100 datasets with 0.1M fewer parameters. It is notable that our HPINet outperforms SwinIR-light by a maximum PSNR
of 0.27dB, which is a significant improvement for image SR.

\subsubsection{Qualitative Comparison}
We further show visual examples of different methods under scaling factor $\times 4$. As shown in  \cref{fig:visual_results1}, our HPINet can
recover more details than IDN, CARN, IMDN, A$^2$F-L and SwinIR-light, which indicates the superiority of our method. 

\begin{figure}[t]
\centering
 \subfloat[w/ GPA]{
     \centering
     \begin{minipage}[b]{0.46\linewidth}
     \includegraphics[width=1\textwidth]{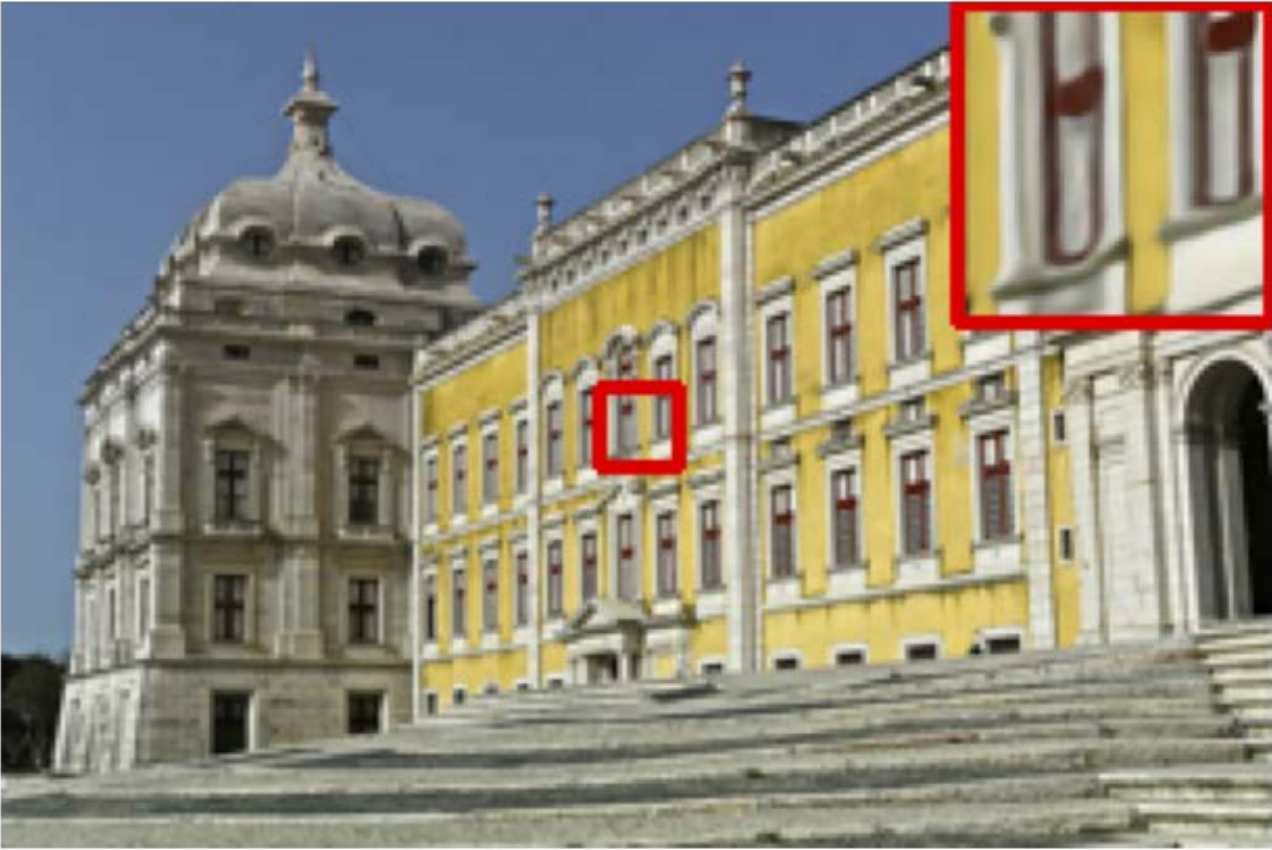}
    \\
     \includegraphics[width=1\textwidth]{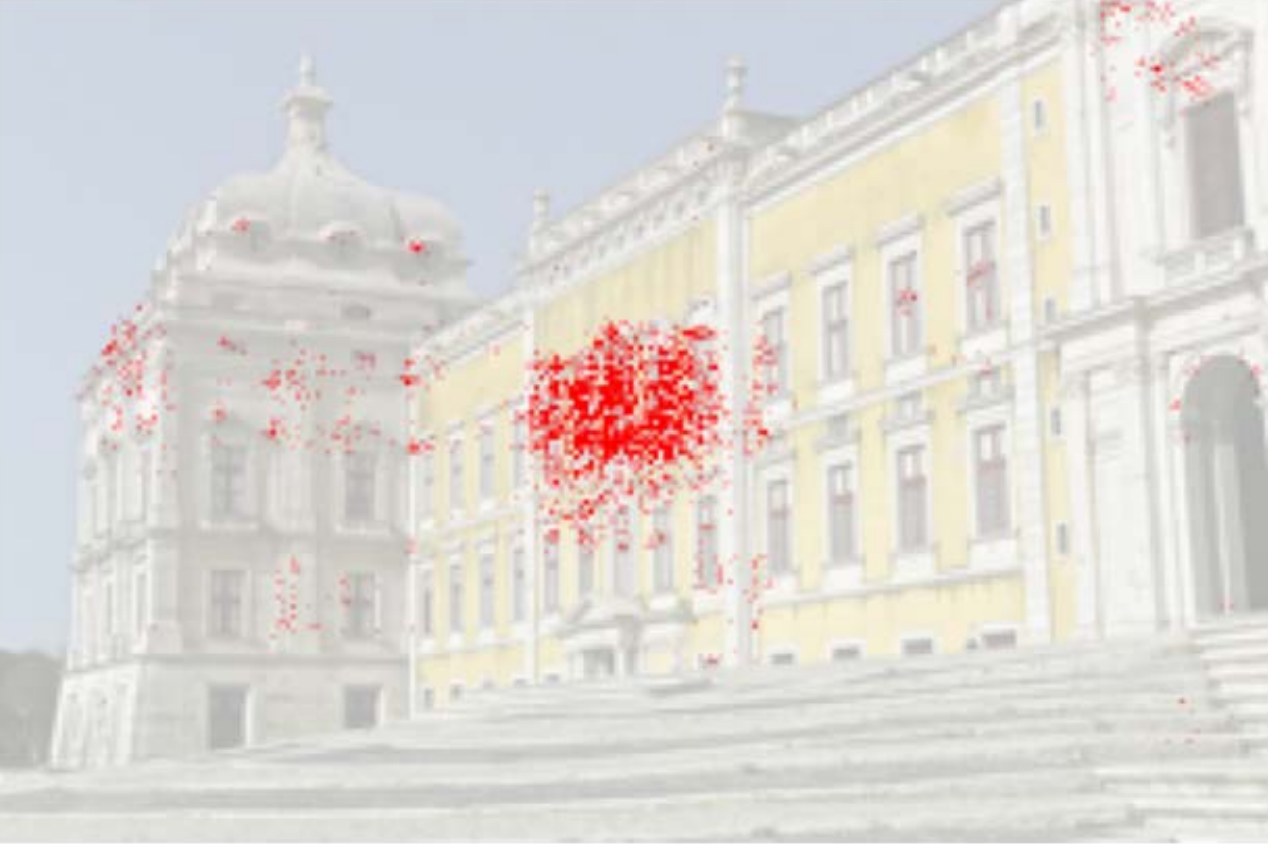}
    \end{minipage}
    }
\subfloat[w/o GPA]{
     \centering
     \begin{minipage}[b]{0.46\linewidth}
     \includegraphics[width=1\textwidth]{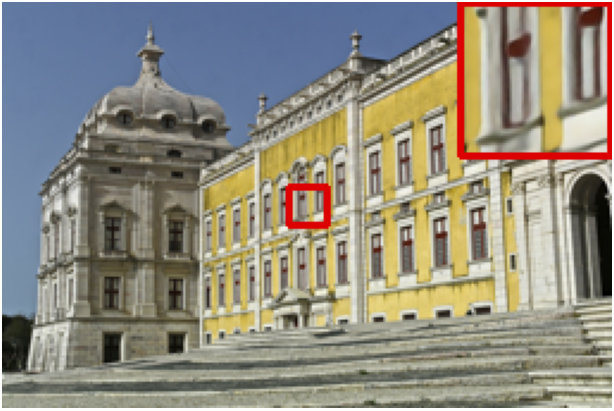}
    \\
     \includegraphics[width=1\textwidth]{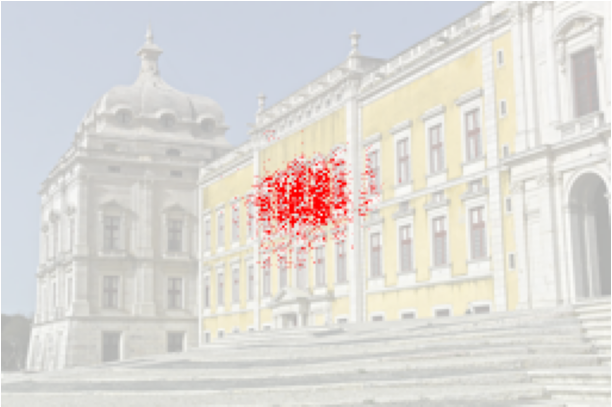}
    \end{minipage}
    }
    \caption{LAM \cite{LAM} comparison between the full HPINet (w/ GPA) and the variant without GPA (w/o GPA) for $\times 4$ SR. The first row shows input along with the predicted result, and the second row shows the effective receptive field. Zoom in for better view.}
\label{fig:lam_gpa}
\end{figure}

\begin{figure}[t]
\centering
\includegraphics[width=0.9\linewidth]{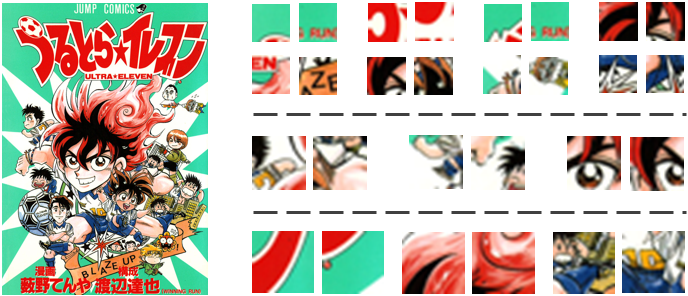}
\caption{Visualization of patch pairs that best match. Patches with different size come from different blocks in the network.}
\label{fig:related}
\end{figure}

\subsection{Ablation Analysis}\label{sec:ablation}

In this section, we conduct ablation experiments to study the effect of Cascaded Patch Division (CPD) and Global Pixel Access (GPA). 
Evaluations are performed on Set5 \cite{set5}, set14 \cite{set14}, B100 \cite{b100}, Urban100 \cite{urban100} and Manga109 \cite{manga109} datasets. Besides PSNR and SSIM, LPIPS \cite{lpips} is adopted to evaluate the perceptual quality of recovered images as well. It can provide better judgment of image quality when two models have similar performances in terms of PSNR and SSIM. We also adopt multiply-accumulate operations (Multi-Adds) on a $1280 \times 720$ query image to evaluate computational complexity. 

We start with a naive baseline by removing both components (model \ding{172}). Then we add CPD (model \ding{173}) and GPA (model \ding{174}) to the baseline, respectively. At last, both components are employed to compose our final version of method (model \ding{175}). The results are reported in \cref{tab:ablation}.
\subsubsection{Effectiveness of CPD}
To show the effect of CPD, we instantiate model \ding{174}  where all blocks share the same patch size for comparison. The patch size is fixed as 18 to maintain similar Multi-Adds. As shown in \cref{tab:ablation}, model \ding{175} improves the performance of model \ding{174} on all metrics. The same phenomena can also be observed by comparing model \ding{173} and model \ding{172}, where the LPIPS gets substantially improvement by adding CPD.
It is notable that the model with CPD turns out to achieve better SSIM and LPIPS, even when its PSNR results are indistinguishable. These results validate that CPD can provide not only higher image similarity but also better perceptual quality.

\subsubsection{Effectiveness of GPA}\label{sec:ablation_gpa}

A core feature of our HPINet is its ability to access global pixel effectively. To highlight the contribution of global modeling, we drop GPA in model \ding{173} for comparison, meanwhile expanding IPSA modules to keep similar parameter budget. Recall that IPSA only integrates pixels within a patch. Comparing model \ding{173} and model \ding{175}, we can observe that with the help of GPA, the PSNR and SSIM get consistent improvements on all five datasets. Specifically, the PSNR increases by a maximum of 0.15dB, which is a notable boost in lightweight image SR. Interestingly, the LPIPS on several benchmark datasets gets worse after adding GPA to the baseline model, and we guess it is caused by the introduction of noisy pixels. This problem can 
be greatly reduced by using the proposed CPD module, which can be observed by comparing model \ding{174} with model \ding{175}.

To better understand the main reason of the improvement brought by GPA, we utilize LAM \cite{LAM} to visualize the effective receptive field of a input patch. As shown in  \cref{fig:lam_gpa}, the patch benefits from a global range of useful pixels by using GPA. In  \cref{fig:related}, we further present examples of patch pairs that matched and selected by GPA in different blocks. It can be found that each patch pair share some visual similarities, which could facilitate restoring more details. All results indicate the effectiveness of the proposed GPA in improving PSNR and SSIM performances.

\begin{table*}
\begin{center}
\caption{Parameter, Running Time and PSNR Comparison for Scale Factor $\times 4$\label{tab:complexity}}
\resizebox{\linewidth}{!}{
\begin{tabular}{lcccccc}
\toprule
Model & Param(M) & \makecell[c]{Set5\\ \hline PSNR(dB)/Time(ms)} & \makecell[c]{Set14\\ \hline PSNR(dB)/Time(ms)} & \makecell[c]{B100\\ \hline PSNR(dB)/Time(ms)}  & \makecell[c]{Urban100\\ \hline PSNR(dB)/Time(ms)}  & \makecell[c]{Manga109\\ \hline PSNR(dB)/Time(ms)}  \\ 
\midrule
RCAN \cite{rcan} & 15.6  & 32.63/56 & 28.87/65 & 27.77/61 & 26.82/119 & 31.22/157 \\ 
SAN \cite{san} & 15.9  & 32.64/68 & 28.92/127 & 27.78/58 & 26.79/997 & 31.18/1771 \\
SwinIR \cite{swinir} & 11.9 & 32.72/65 & 28.94/117 & 27.83/75 & 27.07/401 & 31.67/521 \\ 
\hline
SwinIR-light \cite{swinir} & 0.90 & 32.44/47 & 28.77/59 & 27.69/48 & 26.47/158 & 30.92/198 \\
HPINet-L  & 1.44 & 32.72/39 & 28.97/51 & 27.79/40 & 26.95/144 & 31.47/220 \\ 
HPINet-M  & 0.90 & 32.60/38 & 28.87/49 & 27.73/39 & 26.71/137 & 31.19/194 \\ 
HPINet-S  & 0.46 & 32.47/34 & 28.80/48 & 27.69/36 & 26.59/131 & 30.92/160 \\ 
\bottomrule
\end{tabular}}
\end{center}
\end{table*}

\begin{table}
\begin{center}
    \caption{Ablation Study on IPSA and $3\times 3$ Conv. The PSNR Results on Five Benchmark Datasets Are Included}
\label{tab:ablation2}
\resizebox{\linewidth}{!}{
\begin{tabular}{c|cccccc}
\toprule
Model& Set5 & Set14 & B100 & Urban100 & Manga109\\ 
\midrule
w/o IPSA & 32.26 & 28.66 & 27.60 & 26.30 & 30.66 \\
w/o $3\times 3$ conv & 32.09 & 28.63 & 27.57 & 26.17 & 30.55 \\
HPINet (full) & 32.60 & 28.87 & 27.73 & 26.71 & 31.19 \\
\bottomrule
\end{tabular}}
\end{center}
\end{table}

\subsubsection{Effectiveness of IPSA}\label{sec:ablation_ipsa}
The IPSA module can model long-range dependencies in a patch. To prove the effectiveness of this module, we remove IPSA from all the attention blocks and replace it with depth-wise convolutions to maintain a similar number of parameters. As shown in the first row of \cref{tab:ablation2}, the model “w/o IPSA” behaves much worse than the full HPINet model, which indicates the importance of IPSA in improving the SR performance. Compared with conventional  convolutions, the IPSA can benefit from a wider range of surrounding pixels and thus achieves much higher PSNR values.

\subsubsection{Effectiveness of $3\times3$ Conv}\label{sec:ablation_conv}
The HPI attention block is designed in a hierarchical manner that the tail $3\times3$ convolution is responsible for processing the finest local details. As shown in the second row of \cref{tab:ablation2}, the model “w/o $3\times3$ Conv” achieves much lower PSNR values than the full model, which meets the expectation. It indicates convolution plays a basic role even in Transformer-like models, which is in line with conclusions from other vision task \cite{cvt,cmt,incorporatingcnn}. It is worth emphasizing that convolution is indispensable but not competitive enough. For example, with the equipment of IPSA and GPA, our proposed HPINet surpasses the pure convolution-based model RCAN \cite{rcan} in all datasets with $10\times$ fewer parameters (See \cref{tab:complexity}). Thus it is hard to replace IPSA and GPA with convolution while maintaining similar performance. Together with the aforementioned results, the observation proves effectiveness of our hierarchical design.

\subsection{Model Size and and Running Time Analyses}
To demonstrate the effectiveness and efficiency of HPINet, we design three variants with different model size (S/M/L) and evaluate their PSNR results and speed on five datasets. For simplicity, All variants only differ in number of channels. 

\subsubsection{Model Size} The curve of PSNR \textit{v.s.} model size is depicted in \cref{fig:psnr_size} and the detailed complexity of the
three models are included in \cref{tab:complexity}. We compare our HPINet with other lightweight SR models of various sizes, including MAFFSRN~\cite{maffsrn}, RFDN~\cite{rfdn}, LAPAR-A \cite{lapar}, IMDN~\cite{imdn}, LatticeNet~\cite{latticenet}, SwinIR-light~\cite{swinir}, A$^2$F-L~\cite{a2f}
and A-cubeNet~\cite{acubenet}. Our HPINet-S/M/L achieves much higher PSNR than all other lightweight models at each size. Specially, HPINet-L
outperforms RCAN (15.6M) and SAN (15.9M) with less than 1.5M parameters. 

\subsubsection{Running Time} To reduce the accidental error, we run each model for 10 times on one GPU and calculate the average time as the final running time. We also compare them with other advanced models, including RCAN \cite{rcan}, SAN \cite{san} and SwinIR \cite{swinir}. RCAN \cite{rcan} is a classic CNN-based model and SAN \cite{san} is a often-cited model equipped with non-local modules. SwinIR \cite{swinir} is a state-of-the-art Transformer-based model. According to \cref{tab:complexity}, several observations can be summarized as follows: (1) With the same (\textit{i.e.,} HPINet-M) or fewer parameters (\textit{i.e.,} HPINet-S), our model runs faster than SwinIR-light whiling maintaining higher PSNR values. 
(2) When using a slightly larger model, \textit{i.e.,} HPINet-L, we can even achieve superior or comparable PSNR values with very large models including RCAN, SAN and SiwnIR. (3) Our HPINet runs faster than all other models on Set5, Set14 and B100 datasets. As the image resolution increases on
Urban100 and Manga109 datasets, HPINet is slightly slower than RCAN but much faster than SAN and SwinIR. Overall, the proposed HPINet has a better
trade-off between model complexity and PSNR.

\section{Conclusion}

In this paper, we proposed a lightweight single-image super-resolution network called HPINet which sequentially stacks a series of Hierarchical Pixel Integration (HPI) blocks. The HPI block is designed according to the hierarchical interpretation of a LAM map. Specifically, each block consists of three main components: a Global Pixel Access (GPA) module, an Intra-Patch Self-Attention (IPSA) module and a $3\times3$ convolutional layer. They are responsible for processing input images from coarse to fine. Besides, a Cascaded Patch Division (CPD) strategy is also proposed for better perceptual quality. Benefiting from these components, HPINet can effectively capture the global, long-range and local relations in an efficient manner. Quantitative and qualitative experimental results demonstrate the superior performance of HPINet over previous state-of-the-art SR models on benchmark datasets.  In the future, we will try to improve the perceptual quality of GPA and investigate the potential of HPINet in other low-level tasks.

\section{Acknowledgments}
This work is supported by National Natural Science Foundation of China (No.62076119, No.61921006, No.62072232), Program for Innovative Talents
and Entrepreneur in Jiangsu Province, and Collaborative Innovation Center of Novel Software Technology and Industrialization.
\bigskip

\bibliography{aaai23}

\begin{thebibliography}{51}
\providecommand{\natexlab}[1]{#1}

\bibitem[{Agustsson and Timofte(2017)}]{div2k}
Agustsson, E.; and Timofte, R. 2017.
\newblock Ntire 2017 challenge on single image super-resolution: Dataset and
  study.
\newblock In \emph{Proceedings of the IEEE conference on computer vision and
  pattern recognition workshops}, 126--135.

\bibitem[{Ahn, Kang, and Sohn(2018)}]{carn}
Ahn, N.; Kang, B.; and Sohn, K.-A. 2018.
\newblock Fast, accurate, and lightweight super-resolution with cascading
  residual network.
\newblock In \emph{Proceedings of the European conference on computer vision
  (ECCV)}, 252--268.

\bibitem[{Bevilacqua et~al.(2012)Bevilacqua, Roumy, Guillemot, and
  Morel}]{set5}
Bevilacqua, M.; Roumy, A.; Guillemot, C.; and Morel, M.-L.~A. 2012.
\newblock Low-Complexity Single-Image Super-Resolution based on Nonnegative
  Neighbor Embedding.
\newblock In \emph{British Machine Vision Conference (BMVC)}, 1--10.

\bibitem[{Chen et~al.(2021)Chen, Wang, Guo, Xu, Deng, Liu, Ma, Xu, Xu, and
  Gao}]{ipt}
Chen, H.; Wang, Y.; Guo, T.; Xu, C.; Deng, Y.; Liu, Z.; Ma, S.; Xu, C.; Xu, C.;
  and Gao, W. 2021.
\newblock Pre-trained image processing transformer.
\newblock In \emph{Proceedings of the IEEE/CVF Conference on Computer Vision
  and Pattern Recognition}, 12299--12310.

\bibitem[{Cheng et~al.(2019)Cheng, Matsune, Li, Zhu, Zang, and
  Zhan}]{encoderdecoder}
Cheng, G.; Matsune, A.; Li, Q.; Zhu, L.; Zang, H.; and Zhan, S. 2019.
\newblock Encoder-decoder residual network for real super-resolution.
\newblock In \emph{Proceedings of the IEEE/CVF Conference on Computer Vision
  and Pattern Recognition Workshops}, 0--0.

\bibitem[{Chu et~al.(2021)Chu, Zhang, Ma, Xu, and Li}]{faslr}
Chu, X.; Zhang, B.; Ma, H.; Xu, R.; and Li, Q. 2021.
\newblock Fast, accurate and lightweight super-resolution with neural
  architecture search.
\newblock In \emph{2020 25th International Conference on Pattern Recognition
  (ICPR)}, 59--64.

\bibitem[{Dai et~al.(2019)Dai, Cai, Zhang, Xia, and Zhang}]{san}
Dai, T.; Cai, J.; Zhang, Y.; Xia, S.-T.; and Zhang, L. 2019.
\newblock Second-order attention network for single image super-resolution.
\newblock In \emph{Proceedings of the IEEE/CVF conference on computer vision
  and pattern recognition}, 11065--11074.

\bibitem[{Dong et~al.(2014)Dong, Loy, He, and Tang}]{srcnn}
Dong, C.; Loy, C.~C.; He, K.; and Tang, X. 2014.
\newblock Learning a deep convolutional network for image super-resolution.
\newblock In \emph{European conference on computer vision}, 184--199.

\bibitem[{Dong, Loy, and Tang(2016)}]{fsrcnn}
Dong, C.; Loy, C.~C.; and Tang, X. 2016.
\newblock Accelerating the super-resolution convolutional neural network.
\newblock In \emph{European conference on computer vision}, 391--407.

\bibitem[{Dosovitskiy et~al.(2020)Dosovitskiy, Beyer, Kolesnikov, Weissenborn,
  Zhai, Unterthiner, Dehghani, Minderer, Heigold, Gelly et~al.}]{transformer}
Dosovitskiy, A.; Beyer, L.; Kolesnikov, A.; Weissenborn, D.; Zhai, X.;
  Unterthiner, T.; Dehghani, M.; Minderer, M.; Heigold, G.; Gelly, S.; et~al.
  2020.
\newblock An image is worth 16x16 words: Transformers for image recognition at
  scale.
\newblock \emph{arXiv preprint arXiv:2010.11929}.

\bibitem[{Gu and Dong(2021)}]{LAM}
Gu, J.; and Dong, C. 2021.
\newblock Interpreting super-resolution networks with local attribution maps.
\newblock In \emph{Proceedings of the IEEE/CVF Conference on Computer Vision
  and Pattern Recognition}, 9199--9208.

\bibitem[{Guo et~al.(2022)Guo, Han, Wu, Tang, Chen, Wang, and Xu}]{cmt}
Guo, J.; Han, K.; Wu, H.; Tang, Y.; Chen, X.; Wang, Y.; and Xu, C. 2022.
\newblock Cmt: Convolutional neural networks meet vision transformers.
\newblock In \emph{Proceedings of the IEEE/CVF Conference on Computer Vision
  and Pattern Recognition}, 12175--12185.

\bibitem[{Hang et~al.(2020)Hang, Liao, Yang, Chen, and Zhou}]{acubenet}
Hang, Y.; Liao, Q.; Yang, W.; Chen, Y.; and Zhou, J. 2020.
\newblock Attention cube network for image restoration.
\newblock In \emph{Proceedings of the 28th ACM International Conference on
  Multimedia}, 2562--2570.

\bibitem[{Huang, Singh, and Ahuja(2015)}]{urban100}
Huang, J.-B.; Singh, A.; and Ahuja, N. 2015.
\newblock Single image super-resolution from transformed self-exemplars.
\newblock In \emph{Proceedings of the IEEE conference on computer vision and
  pattern recognition}, 5197--5206.

\bibitem[{Hui et~al.(2019)Hui, Gao, Yang, and Wang}]{imdn}
Hui, Z.; Gao, X.; Yang, Y.; and Wang, X. 2019.
\newblock Lightweight image super-resolution with information
  multi-distillation network.
\newblock In \emph{Proceedings of the 27th acm international conference on
  multimedia}, 2024--2032.

\bibitem[{Hui, Wang, and Gao(2018)}]{idn}
Hui, Z.; Wang, X.; and Gao, X. 2018.
\newblock Fast and accurate single image super-resolution via information
  distillation network.
\newblock In \emph{Proceedings of the IEEE conference on computer vision and
  pattern recognition}, 723--731.

\bibitem[{Jang, Gu, and Poole(2016)}]{gumblesoftmax}
Jang, E.; Gu, S.; and Poole, B. 2016.
\newblock Categorical reparameterization with gumbel-softmax.
\newblock \emph{arXiv preprint arXiv:1611.01144}.

\bibitem[{Kim, Lee, and Lee(2016{\natexlab{a}})}]{vsdr}
Kim, J.; Lee, J.~K.; and Lee, K.~M. 2016{\natexlab{a}}.
\newblock Accurate image super-resolution using very deep convolutional
  networks.
\newblock In \emph{Proceedings of the IEEE conference on computer vision and
  pattern recognition}, 1646--1654.

\bibitem[{Kim, Lee, and Lee(2016{\natexlab{b}})}]{drcn}
Kim, J.; Lee, J.~K.; and Lee, K.~M. 2016{\natexlab{b}}.
\newblock Deeply-recursive convolutional network for image super-resolution.
\newblock In \emph{Proceedings of the IEEE conference on computer vision and
  pattern recognition}, 1637--1645.

\bibitem[{Li et~al.(2020)Li, Zhou, Qi, Jiang, Lu, and Jia}]{lapar}
Li, W.; Zhou, K.; Qi, L.; Jiang, N.; Lu, J.; and Jia, J. 2020.
\newblock Lapar: Linearly-assembled pixel-adaptive regression network for
  single image super-resolution and beyond.
\newblock \emph{Advances in Neural Information Processing Systems}, 33:
  20343--20355.

\bibitem[{Li et~al.(2019)Li, Yang, Liu, Yang, Jeon, and Wu}]{srfbn}
Li, Z.; Yang, J.; Liu, Z.; Yang, X.; Jeon, G.; and Wu, W. 2019.
\newblock Feedback network for image super-resolution.
\newblock In \emph{Proceedings of the IEEE/CVF conference on computer vision
  and pattern recognition}, 3867--3876.

\bibitem[{Liang et~al.(2021)Liang, Cao, Sun, Zhang, Van~Gool, and
  Timofte}]{swinir}
Liang, J.; Cao, J.; Sun, G.; Zhang, K.; Van~Gool, L.; and Timofte, R. 2021.
\newblock Swinir: Image restoration using swin transformer.
\newblock In \emph{Proceedings of the IEEE/CVF International Conference on
  Computer Vision}, 1833--1844.

\bibitem[{Lim et~al.(2017)Lim, Son, Kim, Nah, and Mu~Lee}]{edsr}
Lim, B.; Son, S.; Kim, H.; Nah, S.; and Mu~Lee, K. 2017.
\newblock Enhanced deep residual networks for single image super-resolution.
\newblock In \emph{Proceedings of the IEEE conference on computer vision and
  pattern recognition workshops}, 136--144.

\bibitem[{Liu, Tang, and Wu(2020)}]{rfdn}
Liu, J.; Tang, J.; and Wu, G. 2020.
\newblock Residual feature distillation network for lightweight image
  super-resolution.
\newblock In \emph{European Conference on Computer Vision}, 41--55.

\bibitem[{Liu et~al.(2020)Liu, Zhang, Tang, Tang, and Wu}]{rfanet}
Liu, J.; Zhang, W.; Tang, Y.; Tang, J.; and Wu, G. 2020.
\newblock Residual feature aggregation network for image super-resolution.
\newblock In \emph{Proceedings of the IEEE/CVF conference on computer vision
  and pattern recognition}, 2359--2368.

\bibitem[{Liu et~al.(2018)Liu, Zhang, Zhang, Lin, and Zuo}]{waveletcnn}
Liu, P.; Zhang, H.; Zhang, K.; Lin, L.; and Zuo, W. 2018.
\newblock Multi-level wavelet-CNN for image restoration.
\newblock In \emph{Proceedings of the IEEE conference on computer vision and
  pattern recognition workshops}, 773--782.

\bibitem[{Liu et~al.(2021)Liu, Lin, Cao, Hu, Wei, Zhang, Lin, and
  Guo}]{swintransformer}
Liu, Z.; Lin, Y.; Cao, Y.; Hu, H.; Wei, Y.; Zhang, Z.; Lin, S.; and Guo, B.
  2021.
\newblock Swin transformer: Hierarchical vision transformer using shifted
  windows.
\newblock In \emph{Proceedings of the IEEE/CVF International Conference on
  Computer Vision}, 10012--10022.

\bibitem[{Luo et~al.(2020)Luo, Xie, Zhang, Qu, Li, and Fu}]{latticenet}
Luo, X.; Xie, Y.; Zhang, Y.; Qu, Y.; Li, C.; and Fu, Y. 2020.
\newblock Latticenet: Towards lightweight image super-resolution with lattice
  block.
\newblock In \emph{European Conference on Computer Vision}, 272--289. Springer.

\bibitem[{Mao, Shen, and Yang(2016)}]{verydeepencoder}
Mao, X.; Shen, C.; and Yang, Y.-B. 2016.
\newblock Image restoration using very deep convolutional encoder-decoder
  networks with symmetric skip connections.
\newblock \emph{Advances in neural information processing systems}, 29:
  2802--2810.

\bibitem[{Martin et~al.(2001)Martin, Fowlkes, Tal, and Malik}]{b100}
Martin, D.; Fowlkes, C.; Tal, D.; and Malik, J. 2001.
\newblock A database of human segmented natural images and its application to
  evaluating segmentation algorithms and measuring ecological statistics.
\newblock In \emph{Proceedings Eighth IEEE International Conference on Computer
  Vision. ICCV 2001}, volume~2, 416--425.

\bibitem[{Matsui et~al.(2017)Matsui, Ito, Aramaki, Fujimoto, Ogawa, Yamasaki,
  and Aizawa}]{manga109}
Matsui, Y.; Ito, K.; Aramaki, Y.; Fujimoto, A.; Ogawa, T.; Yamasaki, T.; and
  Aizawa, K. 2017.
\newblock Sketch-based manga retrieval using manga109 dataset.
\newblock \emph{Multimedia Tools and Applications}, 76(20): 21811--21838.

\bibitem[{Muqeet et~al.(2020)Muqeet, Hwang, Yang, Kang, Kim, and Bae}]{maffsrn}
Muqeet, A.; Hwang, J.; Yang, S.; Kang, J.; Kim, Y.; and Bae, S.-H. 2020.
\newblock Multi-attention based ultra lightweight image super-resolution.
\newblock In \emph{European Conference on Computer Vision}, 103--118.

\bibitem[{Niu et~al.(2020)Niu, Wen, Ren, Zhang, Yang, Wang, Zhang, Cao, and
  Shen}]{han}
Niu, B.; Wen, W.; Ren, W.; Zhang, X.; Yang, L.; Wang, S.; Zhang, K.; Cao, X.;
  and Shen, H. 2020.
\newblock Single image super-resolution via a holistic attention network.
\newblock In \emph{European conference on computer vision}, 191--207.

\bibitem[{Shi et~al.(2016)Shi, Caballero, Husz{\'a}r, Totz, Aitken, Bishop,
  Rueckert, and Wang}]{pixelshuffle}
Shi, W.; Caballero, J.; Husz{\'a}r, F.; Totz, J.; Aitken, A.~P.; Bishop, R.;
  Rueckert, D.; and Wang, Z. 2016.
\newblock Real-time single image and video super-resolution using an efficient
  sub-pixel convolutional neural network.
\newblock In \emph{Proceedings of the IEEE conference on computer vision and
  pattern recognition}, 1874--1883.

\bibitem[{Tai, Yang, and Liu(2017)}]{drrn}
Tai, Y.; Yang, J.; and Liu, X. 2017.
\newblock Image super-resolution via deep recursive residual network.
\newblock In \emph{Proceedings of the IEEE conference on computer vision and
  pattern recognition}, 3147--3155.

\bibitem[{Tai et~al.(2017)Tai, Yang, Liu, and Xu}]{memnet}
Tai, Y.; Yang, J.; Liu, X.; and Xu, C. 2017.
\newblock Memnet: A persistent memory network for image restoration.
\newblock In \emph{Proceedings of the IEEE international conference on computer
  vision}, 4539--4547.

\bibitem[{Wang et~al.(2020)Wang, Wang, Zhao, Yan, Fan, and Chen}]{a2f}
Wang, X.; Wang, Q.; Zhao, Y.; Yan, J.; Fan, L.; and Chen, L. 2020.
\newblock Lightweight single-image super-resolution network with attentive
  auxiliary feature learning.
\newblock In \emph{Proceedings of the Asian conference on computer vision},
  volume 12623, 268--285.

\bibitem[{Wang et~al.(2018)Wang, Yu, Wu, Gu, Liu, Dong, Qiao, and
  Change~Loy}]{esrgan}
Wang, X.; Yu, K.; Wu, S.; Gu, J.; Liu, Y.; Dong, C.; Qiao, Y.; and Change~Loy,
  C. 2018.
\newblock Esrgan: Enhanced super-resolution generative adversarial networks.
\newblock In \emph{Proceedings of the European conference on computer vision
  (ECCV) workshops}, 0--0.

\bibitem[{Wang et~al.(2021)Wang, Cun, Bao, and Liu}]{uformer}
Wang, Z.; Cun, X.; Bao, J.; and Liu, J. 2021.
\newblock Uformer: A general u-shaped transformer for image restoration.
\newblock \emph{arXiv preprint arXiv:2106.03106}.

\bibitem[{Wu et~al.(2021)Wu, Xiao, Codella, Liu, Dai, Yuan, and Zhang}]{cvt}
Wu, H.; Xiao, B.; Codella, N.; Liu, M.; Dai, X.; Yuan, L.; and Zhang, L. 2021.
\newblock Cvt: Introducing convolutions to vision transformers.
\newblock In \emph{Proceedings of the IEEE/CVF International Conference on
  Computer Vision}, 22--31.

\bibitem[{Wu et~al.(2020)Wu, Zou, Gui, Zeng, Ye, Zhang, Liu, and Wei}]{mgan}
Wu, H.; Zou, Z.; Gui, J.; Zeng, W.-J.; Ye, J.; Zhang, J.; Liu, H.; and Wei, Z.
  2020.
\newblock Multi-grained attention networks for single image super-resolution.
\newblock \emph{IEEE Transactions on Circuits and Systems for Video
  Technology}, 31(2): 512--522.

\bibitem[{Yuan et~al.(2021)Yuan, Guo, Liu, Zhou, Yu, and Wu}]{incorporatingcnn}
Yuan, K.; Guo, S.; Liu, Z.; Zhou, A.; Yu, F.; and Wu, W. 2021.
\newblock Incorporating convolution designs into visual transformers.
\newblock In \emph{Proceedings of the IEEE/CVF International Conference on
  Computer Vision}, 579--588.

\bibitem[{Zamir et~al.(2021)Zamir, Arora, Khan, Hayat, Khan, and
  Yang}]{restormer}
Zamir, S.~W.; Arora, A.; Khan, S.; Hayat, M.; Khan, F.~S.; and Yang, M.-H.
  2021.
\newblock Restormer: Efficient Transformer for High-Resolution Image
  Restoration.
\newblock \emph{arXiv preprint arXiv:2111.09881}.

\bibitem[{Zeyde, Elad, and Protter(2010)}]{set14}
Zeyde, R.; Elad, M.; and Protter, M. 2010.
\newblock On single image scale-up using sparse-representations.
\newblock In \emph{International conference on curves and surfaces}, 711--730.

\bibitem[{Zhang, Zuo, and Zhang(2018)}]{srmd}
Zhang, K.; Zuo, W.; and Zhang, L. 2018.
\newblock Learning a single convolutional super-resolution network for multiple
  degradations.
\newblock In \emph{Proceedings of the IEEE conference on computer vision and
  pattern recognition}, 3262--3271.

\bibitem[{Zhang et~al.(2018{\natexlab{a}})Zhang, Isola, Efros, Shechtman, and
  Wang}]{lpips}
Zhang, R.; Isola, P.; Efros, A.~A.; Shechtman, E.; and Wang, O.
  2018{\natexlab{a}}.
\newblock The unreasonable effectiveness of deep features as a perceptual
  metric.
\newblock In \emph{Proceedings of the IEEE conference on computer vision and
  pattern recognition}, 586--595.

\bibitem[{Zhang, Zeng, and Zhang(2021)}]{ecbsr}
Zhang, X.; Zeng, H.; and Zhang, L. 2021.
\newblock Edge-oriented convolution block for real-time super resolution on
  mobile devices.
\newblock In \emph{Proceedings of the 29th ACM International Conference on
  Multimedia}, 4034--4043.

\bibitem[{Zhang et~al.(2018{\natexlab{b}})Zhang, Li, Li, Wang, Zhong, and
  Fu}]{rcan}
Zhang, Y.; Li, K.; Li, K.; Wang, L.; Zhong, B.; and Fu, Y. 2018{\natexlab{b}}.
\newblock Image super-resolution using very deep residual channel attention
  networks.
\newblock In \emph{Proceedings of the European conference on computer vision
  (ECCV)}, 286--301.

\bibitem[{Zhang et~al.(2019)Zhang, Li, Li, Zhong, and Fu}]{rnan}
Zhang, Y.; Li, K.; Li, K.; Zhong, B.; and Fu, Y. 2019.
\newblock Residual non-local attention networks for image restoration.
\newblock \emph{arXiv preprint arXiv:1903.10082}.

\bibitem[{Zhang et~al.(2018{\natexlab{c}})Zhang, Tian, Kong, Zhong, and
  Fu}]{rdn}
Zhang, Y.; Tian, Y.; Kong, Y.; Zhong, B.; and Fu, Y. 2018{\natexlab{c}}.
\newblock Residual dense network for image super-resolution.
\newblock In \emph{Proceedings of the IEEE conference on computer vision and
  pattern recognition}, 2472--2481.

\bibitem[{Zhao et~al.(2020)Zhao, Kong, He, Qiao, and Dong}]{pan}
Zhao, H.; Kong, X.; He, J.; Qiao, Y.; and Dong, C. 2020.
\newblock Efficient image super-resolution using pixel attention.
\newblock In \emph{European Conference on Computer Vision}, 56--72.

\end{thebibliography}
\end{document}